\documentclass{ecai} 

\usepackage{latexsym}
\usepackage{amssymb}
\usepackage{amsmath}
\usepackage{amsthm}
\usepackage{booktabs}
\usepackage{enumitem}
\usepackage{graphicx}
\usepackage{color}

\usepackage{bm}
\usepackage{colortbl}
\usepackage{makecell}
\usepackage{multicol}
\usepackage{multirow}
\usepackage{array}
\usepackage{pifont}
\usepackage{float}
\usepackage{adjustbox}
\usepackage{caption}
\usepackage{xcolor}

\newcommand{\xmark}{\ding{55}}
\usepackage[pagebackref,breaklinks,colorlinks,citecolor=green]{hyperref}

\definecolor{mllm_color}{rgb}{0.941,1,0.941}

\makeatletter
\newcommand*{\centerfloat}{%
	\parindent \z@
	\leftskip \z@ \@plus 1fil \@minus \textwidth
	\rightskip\leftskip
	\parfillskip \z@skip}
\makeatother

\usepackage{tcolorbox} 
\tcbuselibrary{listings} 
\newtcblisting{mylisting}{
  arc=3mm,
  colback=red!5!white,
  colframe=red!75!black,
  listing only,
  halign=center,
  valign=center, 
  boxsep=1mm,
  top=1mm,
  bottom=1mm,
  left=1mm,
  right=1mm,
  listing options={basicstyle=\ttfamily\footnotesize, breaklines=true},
}

\newcommand{\BibTeX}{B\kern-.05em{\sc i\kern-.025em b}\kern-.08em\TeX}

\begin{document}


\begin{frontmatter}



\title{u-LLaVA: Unifying Multi-Modal Tasks via Large Language Model}



\author[A]{\fnms{Jinjin}~\snm{Xu}}
\author[A]{\fnms{Liwu}~\snm{Xu}}
\author[A]{\fnms{Yuzhe}~\snm{Yang}} 
\author[A]{\fnms{Xiang}~\snm{Li}} 
\author[A]{\fnms{Fanyi}~\snm{Wang}} 
\author[A]{\fnms{Yanchun}~\snm{Xie}} 
\author[A]{\fnms{Yi-Jie}~\snm{Huang}} 
\author[A]{\fnms{Yaqian}~\snm{Li}\thanks{Corresponding Author. Email: liyaqian@oppo.com.}}

\address[A]{OPPO AI Center} 


\begin{abstract}
Recent advancements in multi-modal large language models (MLLMs) have led to substantial improvements in visual understanding, primarily driven by sophisticated modality alignment strategies. However, predominant approaches prioritize global or regional comprehension, with less focus on fine-grained, pixel-level tasks. To address this gap, we introduce u-LLaVA, an innovative unifying multi-task framework that integrates pixel, regional, and global features to refine the perceptual faculties of MLLMs. We commence by leveraging an efficient modality alignment approach, harnessing both image and video datasets to bolster the model’s foundational understanding across diverse visual contexts. Subsequently, a joint instruction tuning method with task-specific projectors and decoders for end-to-end downstream training is presented. Furthermore, this work contributes a novel mask-based multi-task dataset comprising 277K samples, crafted to challenge and assess the fine-grained perception capabilities of MLLMs. The overall framework is simple, effective, and achieves state-of-the-art performance across multiple benchmarks. We make model, data, and code publicly accessible at \href{https://github.com/OPPOMKLab/u-LLaVA}{https://github.com/OPPOMKLab/u-LLaVA}.
\end{abstract}

\end{frontmatter}


\section{Introduction}

Owing to the intrinsic difficulties associated with feature extraction in computer vision (CV) tasks, researchers have predominantly focused on perception rather than cognition over an extended duration. This emphasis bears a substantial impact on the development and understanding of various CV methodologies \cite{lowe2004distinctive}. Although the development of deep neural networks and pre-training techniques has significantly reduced the difficulty of perception, it remains challenging to achieve homogeneity across downstream tasks due to substantial differences in their respective objectives. Recently, causal large language models such as GPT \cite{radford2018improving,radford2019language,brown2020language}, Gemini \cite{team2023gemini} and  LLaMA \cite{touvron2023llama} have reached or come close to human-level performance on a variety of tasks. These advancements have also motivated researchers to incorporate LLMs as components \cite{liu2024visual,zhu2023minigpt} or core elements \cite{team2023gemini,wang2023cogvlm} in visual tasks, leading to the development of visual language models (VLMs), or multi-modal large language models (MLLMs). As a result, these methods have garnered increasing attention in recent times.

\setlength{\tabcolsep}{5pt}
\renewcommand\arraystretch{1.2}
\begin{table}[t]
\caption{Comparison of comprehension levels supported by existing MLLMs.}
\label{tab:tasks}
\begin{center}
\begin{tabular}{c|ccccc}
\hline
\multirow{2}{*}{Methods}  & \multirow{2}{*}{\makecell{Image}} & \multirow{2}{*}{\makecell{Video}} & \multirow{2}{*}{Region} & \multirow{2}{*}{Pixel} \\
 & & & & & \\
\hline
LLaVA \cite{liu2024visual}       & \checkmark  & \xmark      &  \xmark    &  \xmark   \\
MiniGPT-4 \cite{zhu2023minigpt}       & \checkmark  & \xmark      &  \xmark    &  \xmark   \\
Video-LLaMA \cite{damonlpsg2023videollama}  & \checkmark  & \checkmark   &  \xmark    &  \xmark   \\
Video-ChatGPT \cite{maaz2023video} & \xmark  & \checkmark     &  \xmark    &  \xmark   \\
Shikra \cite{chen2023shikra}       & \checkmark  & \xmark    &  \checkmark    &  \xmark   \\
CogVLM \cite{wang2023cogvlm}       & \checkmark  & \xmark    &  \checkmark    &  \xmark   \\
LISA \cite{lai2023lisa}        & \checkmark  & \xmark      &  \xmark    &  \checkmark   \\
u-LLaVA (ours)      & \checkmark  & \checkmark    &  \checkmark    &  \checkmark   \\ \hline
\end{tabular}
\end{center}
\end{table}

Typically, a multi-modal LLM consists of one or multiple encoders to extract features, paired with suitable mapping components (such as MLP \cite{liu2024visual}, Q-Former\cite{zhu2023minigpt}, or cross-attention \cite{bai2023qwen}), to align the other modalities with the textual domain. In comparison to the impressive performance of MLLMs on general-purpose understanding tasks, such as visual question answering (VQA), their capabilities in regional and pixel-level tasks are somewhat less remarkable \cite{liu2024visual,zhu2023minigpt}. To achieve regional-level understanding, it is usual to convert target coordinates into tokens for causal language modeling, such as Shikra \cite{chen2023shikra} and KOSMOS-2 \cite{peng2023kosmos}. To further realize pixel-level understanding, mask-level decoders or extractors are introduced, such as LISA \cite{lai2023lisa}, Osprey \cite{yuan2023osprey} and Next-Chat \cite{zhang2023next}. However, such region comprehension requires extensive data for training, entailing high costs. Pixel-level understanding methods offer more flexibility, but entail introducing or designing specific segmentation modules.

\begin{figure*}[!htb]
\begin{center}
  \includegraphics[width=2\columnwidth]{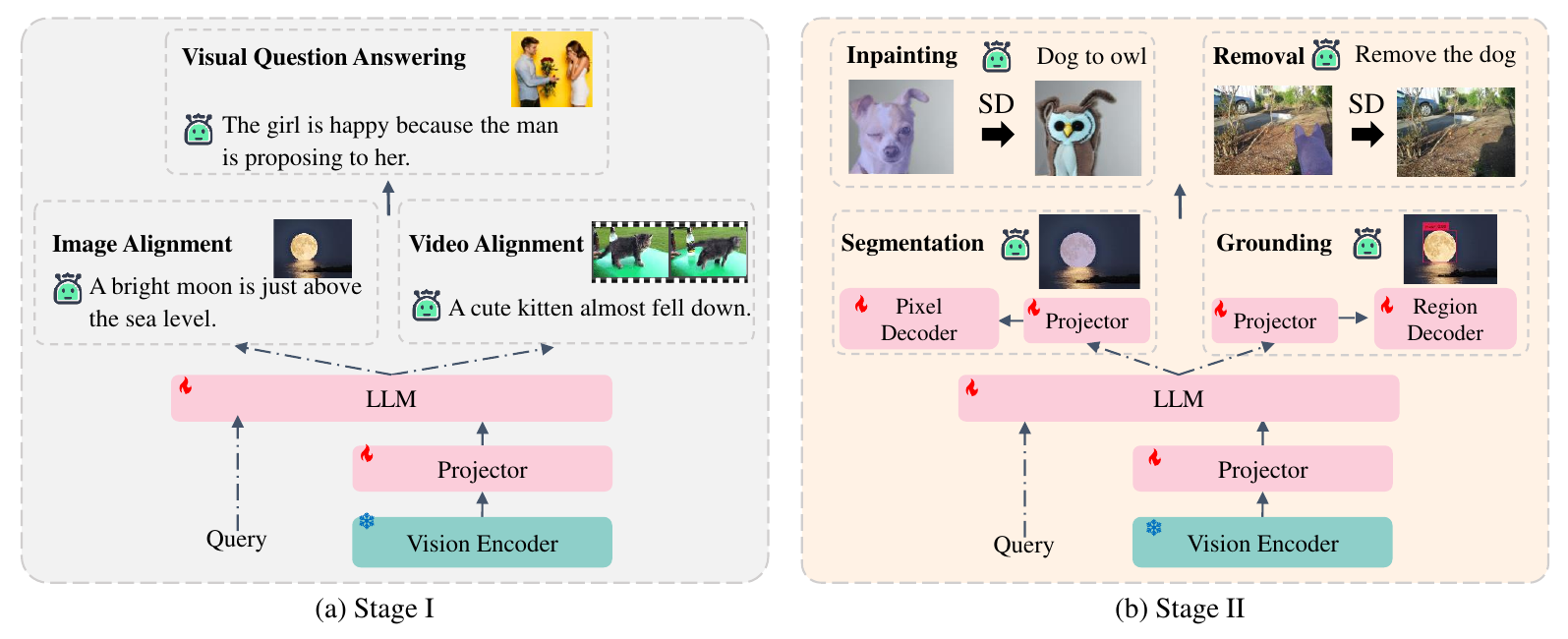}
  \vspace{3mm}
  \caption{Overview of u-LLaVA. In stage I, image and spatio-temporal features are used to efficiently boost the general-purpose modality alignment. In Stage II, task-specific projectors and decoders are jointly trained for region and pixel-level understanding. Further, additional modules such as stable diffusion \cite{rombach2021highresolution} can be easily patched for downstream tasks.}
  \label{fig:pipeline}
\end{center}
\end{figure*}

In this paper, we propose u-LLaVA, a novel approach to enhance the general, region and even pixel-level comprehension abilities of MLLMs. To this end, we first design a efficient visual alignment strategy with image and spatio-temporal representations, and task-specific projectors and decoders are integrated for joint instruction tuning. The overall pipeline is illustrated in Figure \ref{fig:pipeline}.

To enable pixel-level understanding, we employ a projector to connect MLLMs and SAM \cite{lai2023lisa}, achieving two goals: a) imparting semantic understanding capacity to SAM by leveraging the world knowledge inherent in LLM; and b) enhancing the pixel-level understanding ability of LLM by harnessing SAM. To enhance the performance of regional-level comprehension, we introduced a independent location decoder to decode target coordinates from the hidden state or output of MLLMs, which greatly reduces the amount of data required for training. To accommodate the training of the aforementioned models, we have carefully designed a series of task-related prompt pools, and introduced a mask-based, region-specific dataset, namely ullava-277K. Most of the data was collected from publicly available datasets, with missing annotations carefully supplemented by the GPT-3.5.

Contributions can be summarized in three folds:
\begin{itemize}
  \item We propose a efficient visual alignment method for multi-modal pre-training, which leverages image (spatio features) and video (spatio-temporal features) to enhance the perceptual faculties of MLLMs.
  \item We first-time introduce joint instruction tuning approach in the same stage to enable multi-level understanding with task-specific projectors and decoders, see Table \ref{tab:tasks} for details.
  \item We release the joint instruction tuning dataset, ullava-277K, the model, and the code publicly available. Additionally, we conduct comprehensive experiments and demonstrate the effectiveness of the proposed method.
\end{itemize}

\section{Related Work}
\label{sec:formatting}

\subsection{MLLMs}
Surprised by the remarkable abilities of large language models, researchers have shown great interest in transferring the capabilities of LLM to other domains \cite{yin2023survey,wu2023next}. In recent months, remarkable progress has been made in this field, such as LLaVA \cite{liu2024visual}, MiniGPT-4 \cite{zhu2023minigpt}, Otter \cite{li2023otter}, KOSMOS-1/2 \cite{huang2023language,peng2023kosmos}, mPLUG-owl \cite{ye2023mplug} and Flamingo \cite{alayrac2022flamingo}, etc. While having demonstrated impressive performance in image-level understanding, these methods show limited capabilities on pixel or region level tasks.


\subsection{Region-Level Comprehension}
Referring expression comprehension (REC) is one of the most typical region-Level comprehension tasks, and RefCOCO \cite{yu2016modeling}, RefCOCO+ \cite{yu2016modeling} and RefCOCOg \cite{mao2016generation}, RefCLEF \cite{kazemzadeh2014referitgame} are popular datasets for REC. 
Recently, some methods have employed the pix2seq approach to achieve regional understanding \cite{chen2023shikra, peng2023kosmos}. Some strategies further incorporate regional encoding-decoding \cite{zhang2023next}, while others utilize external modules to complete the task \cite{zhao2023bubogpt}.

\subsection{Pixel-Level Understanding}
The advent of MLLMs has reduced the difficulty of subjective visual tasks, but progress on mask-aware tasks, such as referring expression segmentation (RES) and salient object segmentation, has been relatively slow due to the difficulty in designing pixel-level tokens. The prevalent methods currently involve using the output of grounding as the input for SAM \cite{zhang2023next}, or utilizing a specific decoder for end-to-end training \cite{lai2023lisa,yuan2023osprey}.

\section{Methods}

The overall framework of u-LLaVA is presented in Figure \ref{fig:pipeline}. As we can see, u-LLaVA is a multi-modal multitask chatbot that takes text, images, and videos as inputs. It achieves this by unifying the representation space of visual and textual elements at stage I, and understanding region and pixel representations jointly at stage II. In this part, we will first introduce the model architecture and modality alignment strategy in Section \ref{sec:stage1}, followed by a discussion on the joint instruction tuning process in Section \ref{sec:stage2}. Lastly, we will present dataset construction methods.

\subsection{Efficient Visual Alignment}
\label{sec:stage1}

To align representations among different modalities, the projector-based structure is adopted in this work: the pre-trained CLIP ViT-L/14 \cite{radford2021learning} and a visual projector are combined to encode image inputs, while the Vicuna \cite{vicuna2023} is employed as the cognitive module. In addition, u-LLaVA supports video modality by concatenating spatial and temporal representations, requiring only the addition of two special video tokens and a minimal amount of trainable parameters. 

\setlength{\tabcolsep}{1.5pt}
\renewcommand\arraystretch{1.5}
\begin{table}[h]
\begin{center}
\caption{Special tokens for modality and task expressions, where $T$ denotes the number of frames and is set 8 in this work.}
\label{tab:tokens}
\footnotesize
\begin{tabular}{c|ccccc}
\hline
          & Image     & Video   & Tag  & Region & Pixel        \\ \hline
Begin     & \textless{}img\_beg\textgreater{}  & \textless{}vid\_beg\textgreater{}  & \textless{}tag\textgreater{} & \textless{}LOC\textgreater{} & \textless{}SEG\textgreater{} \\
Patches   & \textless{}img\_patch\textgreater{} & \textless{}vid\_patch\textgreater{} & /         & /                   & /                             \\
End       & \textless{}/img\_end\textgreater{}  & \textless{}/vid\_end\textgreater{}  & \textless{}/tag\textgreater{}          & /                 & /   \\ \hline
\makecell{Special token \\ length} & 256     & 256+T          & 1          & 1                            & 1                             \\ \hline
\end{tabular}
\end{center}
\end{table}

Generally, maximizing the likelihood function below to align the representation spaces of image/video and text is a widely-used approach for pre-training \cite{liu2024visual}. For a given image or video embeddings $\bm{x}_e$, and a conversation list of $L$ tokens $\bm{x}_t = \{x_t^1, x_t^2,...,x_t^L\}$, we have the following training objectives, called \textit{\textbf{coarse-grained loss}}:

\begin{equation}
L_{cgl} = \sum_i \log P(x_i | \bm{x}_e, x_{i-k},...,x_{i-1}; \bm{\theta}),
\end{equation}
where, in accordance with \cite{radford2018improving}, $k$, $P$, and $\bm{\theta}$ are the size of context window, the conditional probability, and network parameters, respectively.

\subsection{Joint Instruction Tuning}
\label{sec:stage2}
Visual instruction tuning is a common strategy for MLLM fine-tuning, but most methods only include one or two out of general, region, and pixel-level aspects during the training phase. In this work, we first-time jointly involve general, region, and pixel features in the same tuning stage. 


\textbf{General-aware tuning}: 
In this part, there are no adjustments made to the model structure. However, unlike the first stage, we emphasize the use of multi-turn dialogues and complex reasoning datasets to further enhance the model’s understanding capabilities. The special tokens used in this work are listed in Table \ref{tab:tokens}.

\textbf{Mask-aware tuning}: 
Inspired by LISA \cite{lai2023lisa}, we employ a projector to map the hidden states of the mask-related special tokens and then incorporate them into SAM decoder as the text embeddings to facilitate pixel-level understanding. We use a projector to connect SAM and MLLM, and map the mask-related hidden states as the text embedding for SAM. This endows SAM with semantic perception capabilities while achieving pixel-level perception for MLLM.


\textbf{Region-aware tuning}: 
Similar to pixel perception, we utilize a projector and a location decoder, mapping the hidden states of location-related special tokens directly to the target coordinates, in which the decoder consists of a randomly initialized MLP. To enhance the data volume and improve the decoder's performance, we convert the segmentation annotations into bounding boxes for samples that lack detection annotations.


In general, we have the following training objective, namely \textit{\textbf{fine-grained loss}}:
\begin{equation}
L_{fgl}= L_{cgl} + \left\{
\begin{aligned}
& L_{pixel}, &\text{if mask exists} \\
& L_{region}, &\text{if bbox exists} \\
& 0,  &\text{otherwise}\\
\end{aligned}
\right.
\end{equation}
where the term $L_{pixel} = \alpha_1 L_{bce} + \alpha_2 L_{dice}$ represents the mask prediction loss, and $L_{region} = \beta_1 L_1 + \beta_2 L_{giou}$ is the prediction loss for the target bounding box. The values of $\alpha_1$, $\alpha_2$, $\beta_1$, and $\beta_2$ are set to 2.0, 0.5, 1.0, and 1.0, respectively.


\subsection{Dataset Construction}

To accommodate the training of the aforementioned models, we reorganize or rebuild various types of public datasets, details are summarized in Table \ref{tab:datasets}.

As for the referring and semantic segmentation datasets, all references or semantic labels are extracted and then formed with the given templates. However, salient object detection/segmentation datasets usually lack descriptions of the target objects. To address this issue, we employ mask information to extract the primary objects from images within MSRA-10K \cite{HouPami19Dss} and MSRA-B \cite{WangDRFI2017}. The extracted objects are then fed into BLIP2 \cite{li2023blip} to generate descriptions solely for the objects. Lastly, GPT-4o is used to phrase the object tags from the generated description, followed by the integration of predefined templates to complete the reconstruction process. We refer to the reconstructed salient instruction dataset as Salient-15K for short. The template examples and construction process of Salient-15K are summarized in \hyperref[sec:appendix]{Appendix}. 


\setlength{\tabcolsep}{1pt}
\renewcommand\arraystretch{1.3}
\begin{table}[!t]
\begin{center}
\footnotesize
\caption{Construction of the training datasets. The color \colorbox[rgb]{0.92,0.95,1.0}{blue} indicates that the dataset is utilized in Stage I, while \colorbox[rgb]{0.99,0.96,0.9}{yellow} signifies its usage in Stage II. Where annotations }
\label{tab:datasets}
\begin{tabular}{c|c|c|c}
\hline
\multicolumn{1}{l|}{}         & Dataset    & Images/Videos & Annotations   \\ \hline
\multirow{2}{*}{\makecell{Visual \\ Captioning}} & \cellcolor[HTML]{ECF4FF}{LLaVA CC3M \cite{liu2024visual}}    &\cellcolor[HTML]{ECF4FF}{595,375}    & \cellcolor[HTML]{ECF4FF}{595,375}    \\
                                & \cellcolor[HTML]{ECF4FF}TGIF \cite{li2016tgif}       & \cellcolor[HTML]{ECF4FF}125,782    & \cellcolor[HTML]{ECF4FF}125,782    \\ 
                              \hline
\multirow{5}{*}{VQA} & \cellcolor[HTML]{FDF5E6}{LLaVA-Instruction-mix \cite{liu2024visual}} & \cellcolor[HTML]{FDF5E6}{349,034}  & \cellcolor[HTML]{FDF5E6}{624,610}  \\ 
                            & \cellcolor[HTML]{FDF5E6}{ALLaVA-4V-Instruction \cite{chen2024allava}} & \cellcolor[HTML]{FDF5E6}{643,315}  & \cellcolor[HTML]{FDF5E6}{692,097}  \\ 
                            & \cellcolor[HTML]{FDF5E6}{DVQA \cite{kafle2018dvqa}} & \cellcolor[HTML]{FDF5E6}{10,000}  & \cellcolor[HTML]{FDF5E6}{10,000}  \\ 
                            & \cellcolor[HTML]{FDF5E6}{DocVQA \cite{tito2021document}} & \cellcolor[HTML]{FDF5E6}{9,911}  & \cellcolor[HTML]{FDF5E6}{9,911}  \\ 
                            & \cellcolor[HTML]{FDF5E6}{AI2D \cite{liu2024visual}} & \cellcolor[HTML]{FDF5E6}{4,060}  & \cellcolor[HTML]{FDF5E6}{4,060}  \\ 
    
\hline
\multirow{4}{*}{RES}          
                              & \cellcolor[HTML]{FDF5E6}{RefCOCO \cite{yu2016modeling}}    & \cellcolor[HTML]{FDF5E6}{16,994}      & \cellcolor[HTML]{FDF5E6}{120,624}    \\
                              & \cellcolor[HTML]{FDF5E6}{RefCOCO+ \cite{yu2016modeling}}   & \cellcolor[HTML]{FDF5E6}{16,992}      & \cellcolor[HTML]{FDF5E6}{120,191}    \\
                              & \cellcolor[HTML]{FDF5E6}{RefCOCOg \cite{mao2016generation}}   & \cellcolor[HTML]{FDF5E6}{21,899}      & \cellcolor[HTML]{FDF5E6}{80,512}    \\
                              & \cellcolor[HTML]{FDF5E6}{RefCLEF \cite{kazemzadeh2014referitgame}}  & \cellcolor[HTML]{FDF5E6}{17,978}      & \cellcolor[HTML]{FDF5E6}{108,652}     \\ 
                              \hline
\multirow{4}{*}{\makecell{Semantic \\ Segmentation}} 
                              & \cellcolor[HTML]{FDF5E6}{COCO-Stuff \cite{radford2021learning}} & \cellcolor[HTML]{FDF5E6}{118,205}   & \cellcolor[HTML]{FDF5E6}{742,787}  \\
                              & \cellcolor[HTML]{FDF5E6}{VOC2010 \cite{everingham2010pascal}}    & \cellcolor[HTML]{FDF5E6}{4,366}    & \cellcolor[HTML]{FDF5E6}{81,139}           \\
                              & \cellcolor[HTML]{FDF5E6}{PACO-LVIS \cite{ramanathan2023paco}} & \cellcolor[HTML]{FDF5E6}{45,790} & \cellcolor[HTML]{FDF5E6}{612,188}  \\
                              & \cellcolor[HTML]{FDF5E6}{ADE20K \cite{zhou2017scene}}     & \cellcolor[HTML]{FDF5E6}{20,196}      & \cellcolor[HTML]{FDF5E6}{165,120}           \\ \hline
\multirow{2}{*}{Salient-15K} & \cellcolor[HTML]{FDF5E6}{MSRA-10K \cite{HouPami19Dss}} & \cellcolor[HTML]{FDF5E6}{10,000}  & \cellcolor[HTML]{FDF5E6}{10,000}  \\
                                                    & \cellcolor[HTML]{FDF5E6}{MSRA-B \cite{WangDRFI2017}}  & \cellcolor[HTML]{FDF5E6}{5,000}   & \cellcolor[HTML]{FDF5E6}{5,000} \\
\hline

\end{tabular}
\end{center}
\end{table}

\setlength{\tabcolsep}{7pt}
\renewcommand\arraystretch{1.4}
\begin{table*}[!hbt]
\centerfloat
\caption{RES results with cIoU indicator. Specialists represent models that are specifically designed for CV tasks. Where $^\star$ in MLLMs denotes using LoRA \cite{hu2021lora} for parameter efficient training. The top 2 results
are outlined in \textbf{bold} and with \underline{underline}.}
\label{tab:ciou}
\begin{tabular}{c|l|c|ccc|ccc|cc}
\hline
\multicolumn{1}{c|}{\multirow{2}{*}{Type}} &\multicolumn{1}{c|}{\multirow{2}{*}{Method}} &\multicolumn{1}{c|}{\multirow{2}{*}{\makecell{Segmentation\\Masks}}} & \multicolumn{3}{c|}{RefCOCO} & \multicolumn{3}{c|}{RefCOCO+} & \multicolumn{2}{c}{RefCOCOg} \\ \cline{4-11} 
\multicolumn{1}{c|}{} &\multicolumn{1}{c|}{} & & val    & test A   & test B   & val    & test A    & test B   & val  & test   \\ \hline
\multirow{9}{*}{Specialists} 
& LAVT \cite{Yang_2022_CVPR}      & 0.03M & 72.73 & 75.82 & 68.79 & 62.14 & 68.38 & 55.10 & 61.24 & 62.09   \\
& X-Decoder(L)  \cite{Zou_2023_CVPR} & 0.12M & - & - & - & - & - & - & 64.60 & -  \\ 
& ReLA \cite{Liu_2023_CVPR}  & - & 73.82 & 76.48 & 70.18 & 66.04 & 71.02 & 57.65 & 65.00 & 65.97   \\ 
& SEEM(B) \cite{zou2023segment} & 0.12M & - & - & - & - & - & - & 65.00 & -   \\ 
& SEEM(L) \cite{zou2023segment} & 0.12M & - & - & - & - & - & - & 65.60 & -   \\ 
& PolyFormer(B) \cite{liu2023poly}   & 0.16M & 74.82 & 76.64 & 71.06 & 67.64 & 72.89 & 59.33 & 67.76 & 69.05  \\ 
& PolyFormer(L) \cite{liu2023poly}   & 0.16M & 75.96 & 78.29 & 73.25 & 69.33 & 74.56 & 61.87 & 69.20 & 70.19   \\
& UNINEXT(L) \cite{Yan_2023_CVPR} & 3M & \underline{80.32} & \underline{82.61} & \underline{77.76} & \underline{70.04} & \underline{74.91} & \underline{62.57} & \underline{73.41} & \underline{73.68}   \\ 
& UNINEXT(H) \cite{Yan_2023_CVPR} & 3M & \textbf{82.19} & \textbf{83.44} & \textbf{81.33} & \textbf{72.47} & \textbf{76.42} & \textbf{66.22} & \textbf{74.67} & \textbf{76.37}   \\
\hline
\rowcolor{mllm_color}
& LISA-7B$^\star$ \cite{lai2023lisa}  & $\sim$ 0.80M  & 74.10  & 76.50  & 71.10   & 62.40   & 67.40   & 56.50  & 66.40  & 68.50    \\
\rowcolor{mllm_color} & LISA-7B$^\star$ (ft) \cite{lai2023lisa} & -  & 74.90  & 79.10  & 72.30   & 65.10   & 70.80   & 58.10  & 67.90  & 70.60    \\
\rowcolor{mllm_color} & NExT-Chat-7B \cite{zhang2023next} & 0.15M  & 74.70  & 78.90  & 69.50   & 65.10   & 71.90   & 56.70  & 67.00  & 67.00    \\
\rowcolor{mllm_color} & u-LLaVA-7B$^\star$ (Ours) & $\sim$ 0.66M  & \underline{81.11}  & \underline{82.98}  & \underline{77.63} & \underline{71.36} & \underline{76.88}  & \underline{65.66}  & \underline{73.45}  & \underline{74.65}          \\ 

\rowcolor{mllm_color} \multirow{-5}{*}{MLLMs} & u-LLaVA-7B (Ours) & $\sim$ 0.66M  & \textbf{83.00}  & \textbf{85.09} & \textbf{80.51}  & \textbf{77.10} & \textbf{81.68}  & \textbf{70.56}  & \textbf{77.14} & \textbf{77.97}          \\
\hline
\end{tabular}
\end{table*}

\section{Experiments}

\subsection{Implementation Details}
All experiments are conducted with 8 NVIDIA Tesla A100 80G GPUs and Pytorch framework \cite{NEURIPS2019_9015}. Vicuna v1.1 \cite{vicuna2023} and CLIP ViT-L/14 \cite{radford2021learning} are set to the foundational language model and image encoder. For the region understanding task, the projector and decoder are implemented using two MLPs. Specifically, the projector is configured with layers of [4096->4096, 4096->256], while the decoder comprises layers of [256->256, 256->128, 128->4]. For pixel understanding, a two-layer MLP is used as the projector with layers of [4096->4096, 4096->256]. The decoder is implemented using the off-the-shelf SAM ViT-H \cite{kirillov2023segment}. For alignment and instruction training, AdamW is utilized as the optimizer with a weight decay of 0. The learning rate is set to 2e-3 and 2e-5 (2e-4 if LoRA \cite{hu2021lora} is employed). The batch size per device is configured to 48 and 16 (32 if LoRA), with a gradient accumulation step of 1. Additionally, the token length is set to 1024 and 512. Under the above settings, each training step requires approximately 7s and 5s (9.5s if LoRA), with BF16 and DeepSpeed ZeRO-2 enabled.

\subsection{Evaluation Metrics}
We follow the previous works \cite{Liu_2023_CVPR,lai2023lisa} to validate the quantitative performance of the proposed algorithm, with details as follows:

\textbf{Pixel Segmentation}: Cumulative-IoU (cIoU) is a widely-used performance indicator in segmentation tasks, which calculates the total intersection pixels over the total union pixels. In some works, it is also referred to as the overall-IoU (oIoU), as seen in \cite{Yang_2022_CVPR, Yan_2023_CVPR}.

\textbf{Region Grounding}: The percentage of samples with IoU higher than a threshold X is a commonly used metric in visual grounding tasks, denoted as Precision@X (Prec@X). In this work, we set the threshold to 0.5 according to \cite{chen2023shikra}.

\subsection{Pixel-Level Understanding Performance}
To demonstrate the performance of the proposed method on pixel-level understanding, we conduct experiments on widely-used RES benchmarks, RefCOCO, RefCOCO+, and RefCOCOg. The comparison is made between existing state-of-the-art (SOTA) specialist models and MLLMs with cIoU indicator, as presented in Table \ref{tab:ciou}. 

As can be seen from the table, even with LoRA, our method still achieves the best results among the MLLMs methods. More notably, u-LLaVA-7B surpasses the performance of the prevailing state-of-the-art MLLM method, LISA-7B$^*$(ft), achieving an average improvement of 9.28 in the cIoU indicator. It is also noteworthy that u-LLaVA surpasses the performance of the current leading expert model, UNINEXT(H) \cite{Yan_2023_CVPR}, on the three benchmarks, all while utilizing merely a tenth of the training data. These findings serve as a testament to the efficacy of LLM in tasks that necessitate comprehension-based capabilities.


\subsection{Pixel-level Intent Understanding Performance}
We further examine the zero-shot performance of u-LLaVA in widely recognized salient segmentation datasets to clarify the superiority of MLLMs in comprehending human subjective intentions. 

Here, DUT-OMRON \cite{yang2013saliency} (5,168 test images), DUTS-TE \cite{wang2017learning} (5,019 test images), and ECSSD \cite{shi2015hierarchical} (1000 test images) datasets are selected for validation. To ensure fairness, we draw parallels between our method and a range of other previously conducted unsupervised algorithms. As summarized in Table \ref{tab:salient}, u-LLaVA outperforms the rest, achieving SOTA performance across all three benchmarks, further solidifying the effectiveness and superiority of our method.

\setlength{\tabcolsep}{2pt}
\renewcommand\arraystretch{1.2}
\begin{table}[!t]
\caption{Salient segmentation results on salient object detection benchmarks among different methods, where $\dagger$ denotes the method with Bilateral solver \cite{barron2016fast}, and cIoU is adopted as the metric.}
\label{tab:salient}
\begin{center}
\footnotesize
\begin{tabular}{c|l|c|c|c}
\hline
Type         & \multicolumn{1}{c|}{Method}     & DUT-OMRON  & DUTS-TE   & ECSSD  \\ \hline
\multirow{4}{*}{\makecell{Specialists}}      
& LOST$\dagger$ \cite{simeoni2021localizing} &  48.90  &  57.20  &  72.30 \\
& TokenCut$\dagger$ \cite{wang2022self}           &  61.80  &  62.40  &  77.20 \\
& SELFMASK$\dagger$ \cite{shin2022unsupervised}   &  \textbf{65.50}  &  66.60  &  \textbf{81.80 }\\
& MOVE $\dagger$ \cite{bielski2022move}   &  \underline{63.60}  &  \textbf{68.70}  &  \underline{80.10} \\ 
\hline
\rowcolor{mllm_color} & u-LLaVA-7B$^\star$ (Ours) &  \underline{72.10}   &  \underline{74.64}  &  \underline{89.34} \\
\rowcolor{mllm_color} \multirow{-2}{*}{MLLMs} & u-LLaVA-7B (Ours)         &  \textbf{74.19} &  \textbf{75.97} & \textbf{90.83} \\ \hline
\end{tabular}
\end{center}
\end{table}


\subsection{Region-Level Understanding Performance}
In this section, we conduct a comparative analysis to evaluate u-LLaVA’s performance against other 7B MLLM models in the context of region-level understanding tasks, using the REC task as the benchmark.

\begin{figure}[htb]
\begin{center}
  \includegraphics[width=1\columnwidth]{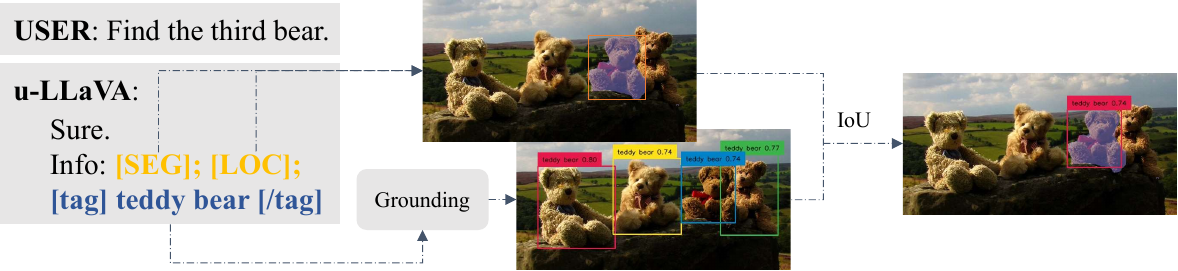}
  \vspace{3mm}
  \caption{The inference workflow of u-LLaVA for region-level comprehension tasks.}
  \label{fig:rec}
\end{center}
\end{figure}

It should be highlighted that we incorporate all intermediate results of our model, including the output of the region decoder and the generated mask, to generate the final regression box thus optimizing the performance. One can further utilize the off-the-shelf grounding model and the generated tags for redundancy, as encapsulated in Figure \ref{fig:rec}, and the corresponding experimental results are summarized in Table \ref{tab:grounding}. Observably, u-LLaVA outperforms other MLLMs such as Shikra \cite{chen2023shikra}, while utilizing a mere one-tenth of the data. However, there exists a discernible performance gap when compared to expert models such as UNINEXT(H), it is essential to consider that this is influenced by multiple factors, including, but not limited to, input resolution and task interference.

\setlength{\tabcolsep}{7pt}
\renewcommand\arraystretch{1.3}
\begin{table*}
\centerfloat
\begin{minipage}{1\linewidth}
\caption{Comparative experiments of existing 7B MLLM models on REC tasks with Prec@0.5 indicator. The top 2 results
are outlined in \textbf{bold} and with \underline{underline}.}
\label{tab:grounding}
\end{minipage}
\begin{adjustbox}{center}
\begin{tabular}{c|l|c|ccc|ccc|cc}
\hline
\multirow{2}{*}{Type}  & \multicolumn{1}{c|}{\multirow{2}{*}{Method}}  &\multirow{2}{*}{\makecell{Grounding\\Boxes}} & \multicolumn{3}{c|}{RefCOCO} & \multicolumn{3}{c|}{RefCOCO+} & \multicolumn{2}{c}{RefCOCOg} \\ \cline{4-11} 

&  & & val & test A  & test B  & val  & test A   & test B  & val  & test \\ \hline

\multirow{5}{*}{Specialists}                              
& SeqTR \cite{zhu2022seqtr} & 1.5M & 87.00  & 90.15  & 83.59 & 78.69 & 84.51  & 71.87 & 82.69  & 83.37   \\
& GroundingDINO(L) \cite{liu2023grounding} & - & 90.56  & 93.19  & 88.24 & 82.75 & 88.95  & 75.92 & 86.13  & 87.02      \\ 
& OFA \cite{wang2022ofa} & - & \underline{92.04}  & \underline{94.03}  & 88.44 & \textbf{87.86} & \textbf{91.70}  & \textbf{80.71} & \underline{88.07}  & \underline{88.78}   \\
& UNINEXT(L) \cite{Yan_2023_CVPR} & 3M & 91.43  & 93.73 & \underline{88.93} & 83.09 & 87.90  & 76.15 & 86.91  & 87.48   \\
& UNINEXT(H) \cite{Yan_2023_CVPR} & 3M & \textbf{92.64}  & \textbf{94.33} & \textbf{91.46} & \underline{85.24} & \underline{89.63}  & \underline{79.79} & \textbf{88.73}  & \textbf{89.37}   \\
\hline

\rowcolor{mllm_color} & Shikra-7B \cite{chen2023shikra}  & $\sim$ 4M   & \underline{87.01}  & \underline{90.61}  & \underline{80.24}  & \underline{81.60} & \underline{87.36}  & \underline{72.12} & \underline{82.27}  & \underline{82.19}   \\

\rowcolor{mllm_color} & VisonLLM-H-7B \cite{wang2024visionllm}  & 0.15M  & -  & 86.70  & -  & - & -  & - & -  & -    \\

\rowcolor{mllm_color} & NeXT-Chat-7B \cite{zhang2023next}  & $\sim$ 4M   & 85.50  & 90.00  & 77.90  & 77.20 & 84.50  & 68.00 & 80.10  & 79.80    \\

\rowcolor{mllm_color} & u-LLaVA-7B$^*$ (Ours) & 0.66M   & 82.95 & 89.08 & 76.29 & 72.91 & 82.43  & 63.41 & 76.23  & 76.56    \\

\rowcolor{mllm_color} \multirow{-5}{*}{MLLMs} & u-LLaVA-7B (Ours) & 0.66M   & \textbf{91.20}  & \textbf{94.29}  & \textbf{87.22}  & \textbf{85.48} & \textbf{90.76}  & \textbf{78.11} & \textbf{86.54}  & \textbf{87.25}    \\

\hline

\end{tabular}
\end{adjustbox}
\end{table*}

\begin{figure*}[!htb]
\begin{center}
  \includegraphics[width=1.9\columnwidth]{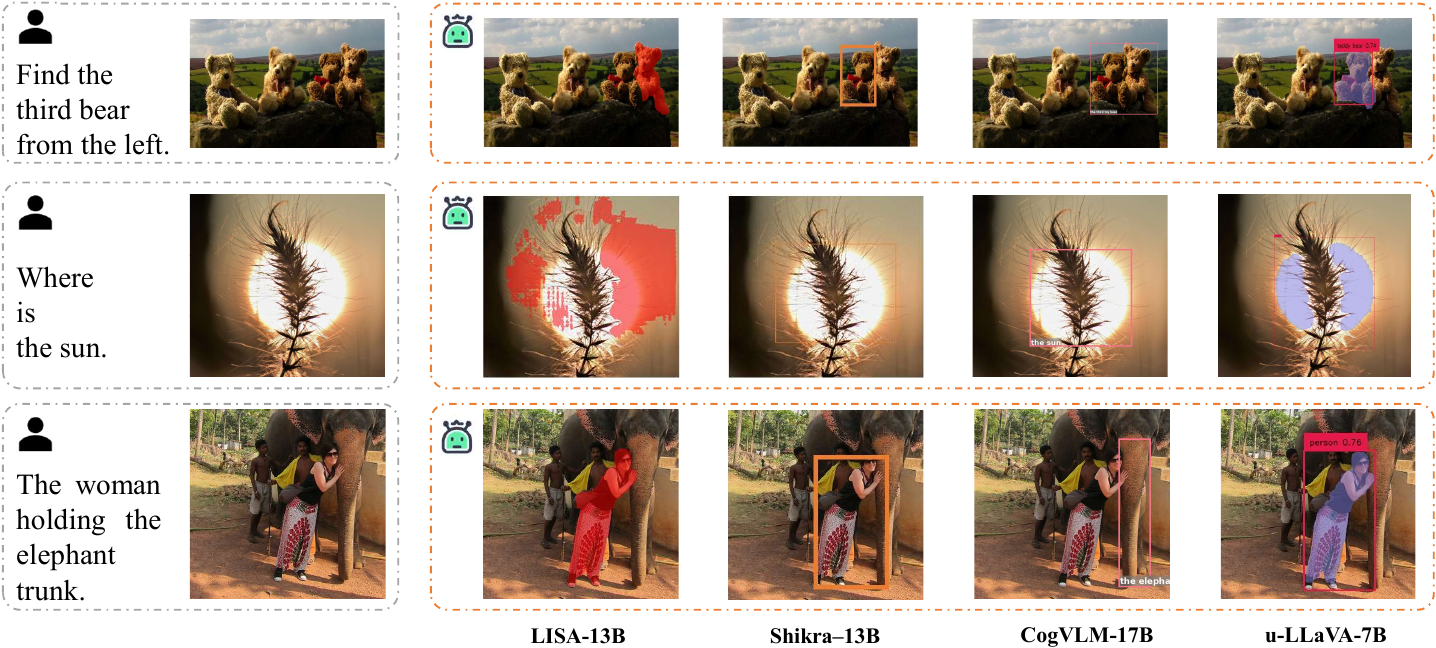}
  \vspace{3mm}
  \caption{Qualitative examples of existing methods for regional and pixel-level understanding.}
  \label{fig:compare}
\end{center}
\end{figure*}

\subsection{General Benchmarks}

In Table \ref{tab:benchmarks}, we present a comparison of our model, u-LLaVA, with popular 7B MLLMs across several multi-modal benchmarks, including MMBench-Dev/Test \cite{liu2023mmbench}, TextVQA \cite{singh2019towards}, GQA \cite{hudson2019gqa}, ScienceQA-IMG \cite{lu2022learn}, and RefCOCO val. Notably, we enlarge the input resolution of u-LLaVA to 336, namely u-LLaVA-1.5, to enhance the performance of the model on such tasks. While these tasks are not the primary focus of the present study, our method demonstrates competitive performance relative to other 7B models. Specifically, u-LLaVA-1.5 achieves best results on the ScienceQA-IMG task using Vicuna-7B-v1.1 and ranks second only to LLaVA-1.5 on the MMBench-Test and GQA benchmarks.

\setlength{\tabcolsep}{1.8pt}
\renewcommand\arraystretch{1.3}
\begin{table*}[htb]
\centerfloat
\begin{minipage}{1\linewidth}
\caption{Experimental results with leading methods on popular multi-modal benchmarks, where the symbol $\dagger$ indicates that the model is trained for 1 epoch for fair comparison. The top 2 results
are outlined in \textbf{bold} and with \underline{underline}.}
\label{tab:benchmarks}
\end{minipage}
\begin{tabular}{llcc|cc|ccc|c|c}
\hline
\multicolumn{1}{c}{\multirow{2}{*}{Method}} & \multicolumn{1}{c}{\multirow{2}{*}{LLM}} & \multicolumn{1}{c}{\multirow{2}{*}{Epoch}} & \multirow{2}{*}{\begin{tabular}[c]{@{}c@{}}Image\\ Size\end{tabular}} & \multicolumn{2}{c|}{General}   & \multicolumn{3}{c|}{VQA} & RES  & REC  \\ \cline{5-11} 
\multicolumn{1}{c}{}      & \multicolumn{1}{c}{}  & \multicolumn{1}{c}{}    &           & MMB$^D$ & MMB$^T$  & TextVQA   & GQA & SciQA$^{IMG}$  & RefCOCO & RefCOCO  \\ \hline
InstructBLIP \cite{liu2024visual}  & Vicuna-7B  &- & 224  & -    & 36      &  -    & -   & - & - & -   \\
Shikra \cite{chen2023shikra}  & Vicuna-7B  & 3 & 224  & -    & -      &  -    & -   & - & - & 87.01   \\
IDEFICS-9B \cite{laurenccon2024obelics} & Vicuna-7B   & - & 224      & 48.2 & 45.3 &  25.9   &  38.4  &    -  & - & -   \\
Qwen-VL \cite{bai2023qwen}  & Qwen-7B   & - & 448      & 38.2   & 32.2   &   \underline{63.8 }&  -  &  \underline{67.1}  & - & 89.4  \\
Mini-Gemini \cite{li2024mini}  & Vicuna-7B-v.5   & - & 336(768)      & \textbf{69.3} & - &  \textbf{65.2}   &  - & - & - & -  \\
LLaVA \cite{liu2024visual}   & Vicuna-7B-v1.1 & 1  & 224     & 38.7    & - &   -     &  -  &  - & - & - \\
LLaVA-1.5 \cite{liu2024improved}   & Vicuna-7B-v1.5 & 1 & 336   & \underline{65.2}    &  \textbf{66.5}       & 58.5   &  \textbf{62.0} & 66.8   & - & -  \\ \hline
\rowcolor{mllm_color} u-LLaVA $\dagger$  & Vicuna-7B-v1.1  & 1 & 224     & 56.2    & 57.0       &  47.4   &  55.8  & 64.5 & 80.01 & 83.96  \\ 
\rowcolor{mllm_color} u-LLaVA-1.5 $\dagger$  & Vicuna-7B-v1.1 & 1   & 336    &  61.6   & \underline{62.3}   &  55.6   &  \underline{57.8}  & \textbf{67.2} & \textbf{81.05} & 86.81  \\ \hline
\end{tabular}
\end{table*}

\subsection{Dataset Ablation}
As shown in Table \ref{tab:ablation_dataset}, we validate the impact of employing varied types of datasets during the second stage of the model’s training on its overall performance. The results indicate that embracing diversity in dataset types fosters improved generalization of the algorithm, thereby circumventing the potential risk of overfitting on specific tasks. In essence, the algorithm’s robustness is enhanced with an increased variety of dataset types.

\setlength{\tabcolsep}{2pt}
\renewcommand\arraystretch{2}
\begin{table}[htb]
\begin{center}
\caption{Ablations on the Stage II training datasets, and cIoU is used as the performance indicator.}
\scriptsize
\label{tab:ablation_dataset}
\begin{tabular}{c|cccc|cc}
\hline

\multirow{2}{*}{Exp.} & \multirow{2}{*}{Referring} & \multirow{2}{*}{Semantic}  & \multirow{2}{*}{Salient}  &\multirow{2}{*}{VQA} & \multirow{2}{*}{\makecell{RefCOCOg\\test}} & \multirow{2}{*}{\makecell{DUT-\\OMRON}} \\ 
&           &               &          &               &             &       \\ \hline

1    & \checkmark  &            &                         &       & 72.83      &  52.04    \\
2    & \checkmark  & \checkmark &                         &       & 75.04      &  46.70    \\
3    & \checkmark  & \checkmark & \checkmark           &       & \underline{75.10}     &  \underline{65.45}    \\
4    & \checkmark  & \checkmark & \checkmark   & \checkmark     & \textbf{77.97}   & \textbf{74.19} \\ \hline
\end{tabular}
\end{center}
\end{table}

\subsection{Qualitative examples}
 
\begin{figure*}[!ht]
\begin{center}
  \includegraphics[width=1.9\columnwidth]{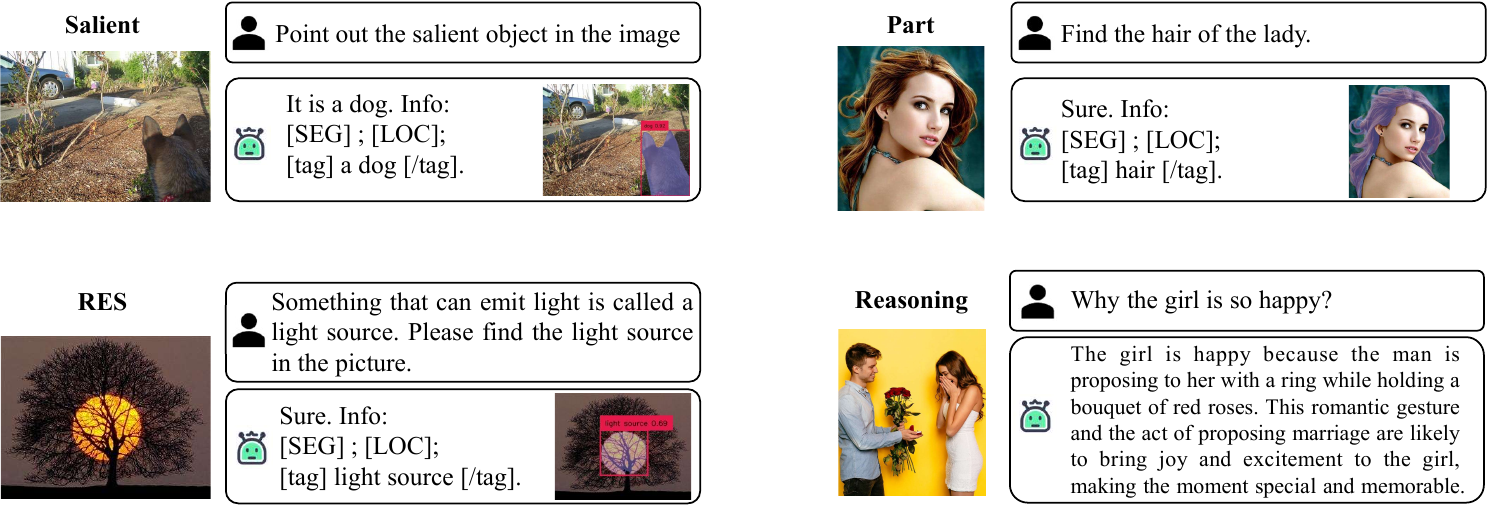}
  \vspace{3mm}
  \caption{Conversation examples with u-LLaVA-7B.}
  \label{fig:exp2}
\end{center}

\end{figure*}

\begin{figure*}[!ht]
\begin{center}
  \includegraphics[width=1.9\columnwidth]{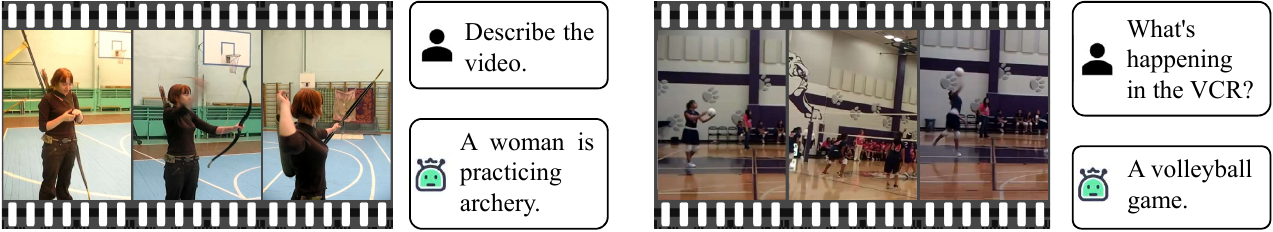}
  \vspace{3mm}
  \caption{Video captioning examples with u-LLaVA-7B.}
  \label{fig:exp3}
\end{center}
\end{figure*}

Qualitative comparison with existing multi-task MLLM methods, LISA \cite{lai2023lisa}, Shikra \cite{chen2023shikra} and CogVLM \cite{wang2023cogvlm}, on grounding and segmentation tasks are given in Figure \ref{fig:compare}. More conversation illustrations can be found in Figure \ref{fig:exp2} and Figure \ref{fig:exp3}.


\begin{figure*}[!ht]
\begin{center}
  \includegraphics[width=1.9\columnwidth]{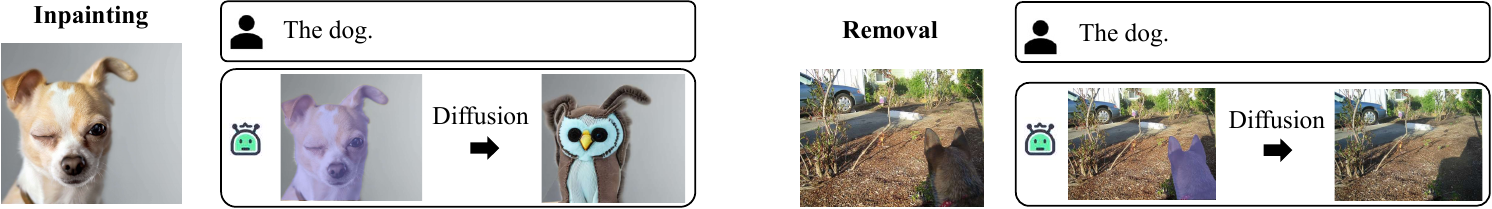}
  \vspace{3mm}
  \caption{Image inpainting and object removal examples with u-LLaVA-7B.}
  \label{fig:exp4}
\end{center}
\end{figure*}

\section{Conclusions}

In this work, we introduce u-LLaVA, a multi-modal large language model that jointly tunes instructions at the global, regional, and pixel levels. Through innovative structural design and data configuration, we have achieved optimal performance in various comprehension-based tasks.

Currently, the pre-training and task adaptation of MLLMs remain an open area with many directions yet to be explored. This study represents an exploratory and experimental effort building upon previous works such as LLaVA and LISA. We believe that the open-sourcing of our work can provide valuable assistance to the development of this field.

\section{Acknowledgement}
This work is sponsored by Shanghai Pujiang Program (23PJ1421800).



\section{Appendix}
\label{sec:appendix}

\subsection{Templates}
Here, we present examples of task templates used by u-LLaVA on different type of training data.

\paragraph{Template examples for salient segmentation task} \quad

\begin{mylisting}
<image> What makes the image stand out?
<image> What is salient one in this image?
<image> Look at the image, segment the main object in the picture and explain.

\end{mylisting}

\paragraph{Template examples for video captioning task} \quad

\begin{mylisting}
<video> Describe the video concisely.
<video> What's happening in this video?
<video> Write a terse but informative summary of the VCR.

\end{mylisting}

\paragraph{Template examples for RES task} \quad

\begin{mylisting}
<image> Segment out the <class>.
<image> Output the mask of the <class>.
<image> Find the <class> in the picture.
\end{mylisting}

\subsection{Construction of Salient-15K}
\begin{figure}[ht]
\begin{center}
  \includegraphics[width=1\columnwidth]{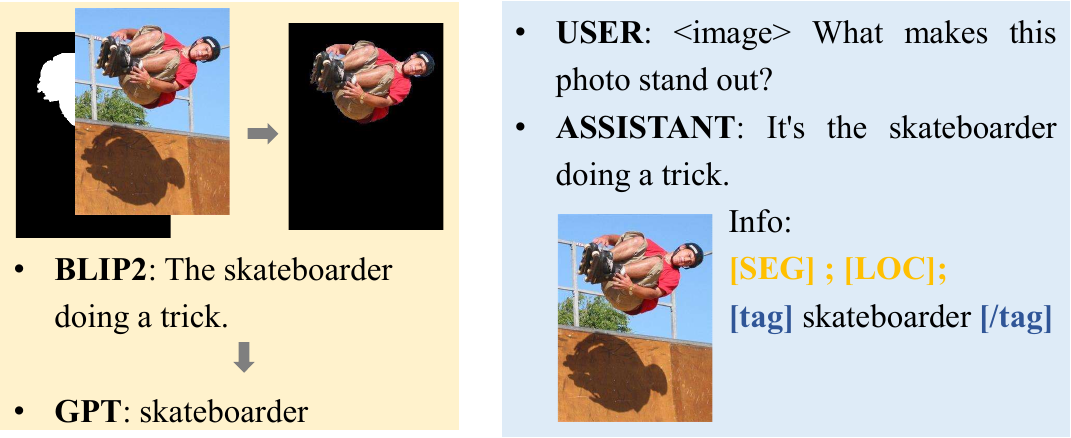}
  \vspace{3mm}
  \caption{The process workflow of Salient-15K.}
  \label{fig:msra_construct}
\end{center}
\end{figure}

As shown in Figure \ref{fig:msra_construct}, given that the MSRA-10K and MSRA-B datasets are devoid of label information and image descriptions pertaining to the principal subjects, we initially proceed by extracting the subjects from the images and subsequently inputting them into BLIP2 \cite{li2023blip} for a rudimentary description. Following this, we employ GPT3.5 to parse the target labels emanating from the elementary description, thereby enabling an expansion of the description information. This approach facilitates a more comprehensive understanding of the subjects within the datasets while compensating for the initial lack of descriptive data.

\bibliography{main}

\end{document}